\documentclass[letter, 10pt, conference]{./ieeeconf}
\makeatletter
\def\endthebibliography{%
  \def\@noitemerr{\@latex@warning{Empty `thebibliography' environment}}%
  \endlist
}
\makeatother

\IEEEoverridecommandlockouts 
\usepackage{times} 
\usepackage{enumerate} 
\usepackage{amsmath} 
\usepackage{mathrsfs}  
\usepackage{bm}
\usepackage{amssymb}  
\usepackage{xcolor}
\usepackage{optidef}
\usepackage{glossaries}
\usepackage{booktabs}
\usepackage[font=footnotesize]{caption}
\usepackage{subcaption}
\usepackage{algpseudocode}

\usepackage[compress]{cite}

\makeatletter
\let\NAT@parse\undefined
\makeatother
\usepackage[pagebackref=false,breaklinks=true,colorlinks=true,bookmarks=false,allcolors=black,citecolor=black]{hyperref}

\usepackage{graphicx}
\graphicspath{{figures/}}
\usepackage{cleveref}
\Crefname{figure}{Fig.}{Figs.}
\Crefname{equation}{Eq.}{Eqs.}
\usepackage{physics}
\usepackage[mode=text]{siunitx}
\usepackage{todonotes}

\usepackage{tikz}

\usepackage[utf8]{inputenc}
\usepackage[T1]{fontenc}
\usepackage[USenglish]{babel}

\newcommand{\eg}{\textit{e.g.}}

\newcommand{\R}{\mathbb{R}}
\newcommand{\Z}{\mathbb{Z}}
\newcommand{\N}{\mathbb{N}}

\DeclareMathOperator{\wrap}{wrap}

\newcommand\copyrighttext{%
  \footnotesize © 2023 IEEE.  Personal use of this material is permitted.  Permission from IEEE must be obtained for all other uses, in any current or future media, including reprinting/republishing this material for advertising or promotional purposes, creating new collective works, for resale or redistribution to servers or lists, or reuse of any copyrighted component of this work in other works.}
\newcommand\copyrightnotice{%
\begin{tikzpicture}[overlay, remember picture]
\node[anchor=south,yshift=10pt] at (current page.south) {\fbox{\parbox{\dimexpr\textwidth-\fboxsep-\fboxrule\relax}{\copyrighttext}}};
\end{tikzpicture}%
}

\usepackage{amsthm}

\theoremstyle{plain}

\newtheorem{remark}{\bf{Remark}}

\newcommand{\Ordo}{\mathcal{O}}
\renewcommand{\vec}[1]{\boldsymbol{\mathbf{#1}}}

\newcommand{\sigmaAng}{\ensuremath{\sigma_\mathrm{ang}}}
\newcommand{\sigmaPos}{\ensuremath{\sigma_\mathrm{pos}}}

\newcommand{\Mthreefivehundred}{\texttt{M3500}}
\newcommand{\sphere}{\texttt{sphere}}
\newcommand{\torus}{\texttt{torus}}
\newcommand{\cube}{\texttt{cube}}
\newcommand{\cubicle}{\texttt{cubicle}}
\newcommand{\rim}{\texttt{rim}}
\newcommand{\MultiThreeFivehundredXeight}{\texttt{Multi3500x8}}
\newcommand{\MultiTenThousandXfive}{\texttt{Multi10000x5}}

\IEEEtriggeratref{11}

\usepackage{listings}
\makeatletter
\def\lst@makecaption{%
  \def\@captype{table}%
  \@makecaption
}
\makeatother

\title{\LARGE \bf
kollagen: A Collaborative SLAM Pose Graph Generator
}

\author{Roberto C. Sundin\textsuperscript{} and David Umsonst\textsuperscript{}
\thanks{\textsuperscript{}Roberto C. Sundin and David Umsonst are with Ericsson Research, Stockholm, Sweden.
        E-Mail: {\tt\small \{roberto.castro.sundin, david.umsonst\}@ericsson.com}}
}

\makeatletter
\def\footnoterule{\relax%
  \kern-5pt
  \hbox to \columnwidth{\vrule width 0.5\columnwidth height 0.4pt\hfill}
  \kern4.6pt}
\makeatother

\begin{document}

\maketitle
\thispagestyle{empty}
\pagestyle{empty}

\begin{abstract}
    In this paper, we address the lack of datasets for -- and the issue
    of reproducibility in -- collaborative SLAM pose graph optimizers by providing a novel pose graph generator.
    Our pose graph generator, \emph{kollagen}, is based on a
    random walk in a planar grid world, similar to the popular \Mthreefivehundred{} dataset for single agent SLAM.
    It is simple to use and the user can set several parameters, \eg, the number of agents, the number of nodes, loop closure generation probabilities, and standard deviations of the measurement noise.
    Furthermore, a qualitative execution time analysis of our pose graph generator showcases the speed of the generator in the tunable parameters.
    
    In addition to the pose graph generator, our paper provides two example
    datasets that researchers can use out-of-the-box to evaluate their algorithms. 
    One of the datasets has 8 agents, each with 3500 nodes, and 67645 constraints in the pose graphs, while the other has 5 agents, each with 10000 nodes, and 76134 constraints. 
    In addition, we show that current state-of-the-art pose graph optimizers are able to process our generated datasets and perform pose graph optimization.
    
    The data generator can be found at \url{https://github.com/EricssonResearch/kollagen}.
\end{abstract}
\section{Introduction}\copyrightnotice
Both robots and extended reality devices, such as AR/VR glasses, permeate the public and private sector more and more in the form of, \eg, industrial applications and consumer devices.
For an autonomous operation, the robots and devices often need to build a map of the environment by themselves and orient themselves in said map.
For this, Simultaneous Localization and Mapping (SLAM) is used \cite{SLAMSurvey}. 
Due to the sheer amount of deployed devices, the large scale environments, or both, it is deemed beneficial to perform collaborative SLAM (C-SLAM) among the devices \cite{CollabSLAMSurvey}.
Therefore, {C-SLAM} has become an active research area in recent years.

In C-SLAM, each agent has a front-end and a back-end. 
The front-end processes measurements and generates loop closures, which indicate correspondences between places that the agent has visited. 
Loop closures are not only made between poses of a single agent as in SLAM (called \emph{intra-agent loop closures}), but also between poses of different agents (called \emph{inter-agent loop closures}). 
Since the front-end of an agent is highly dependent on the types of sensors used by the agent, we put our focus on the back-end in this paper.
In the back-end, the map is abstracted into a graph and graph optimization is performed to optimize over the agent's pose and the position of landmarks.
The graphs of different agents will be connected via inter-agent loop closures.
Often, the landmarks are marginalized out, such that the graph representing the map is a \emph{pose graph}.
The nodes of the pose graph represent the poses of an agent, while the edges represent constraints coming, \eg, from odometry and loop closures.
For SLAM, pose graph optimizers have been developed that are able to solve the pose graph optimization problem of the back-end in an incremental manner, such as iSAM2 \cite{iSAM2}, or provide a certifiably correct solution in an offline manner such as SE-Sync \cite{se-sync}.
In addition to this, there are many datasets for pose graph optimization in the single agent case available, such as \Mthreefivehundred{} \cite{M3500}, and \torus{} and \rim{} \cite{Carlone_2015}. 
There are also several methods available for pose graph optimization in the C-SLAM case. For example, DPGO \cite{DistributedPGO} presents a distributed way to obtain a certifiably correct solution to the collaborative (distributed) pose graph optimization problem, while a majorization minimization approach is proposed in \cite{MajorMinPoseGraph}.
Since the increased interest in SLAM led to an increase in SLAM datasets, one would expect that the increased interest in C-SLAM would lead to an increase in datasets for C-SLAM.
However, to the best of the authors' knowledge, this has not been the case. 
Currently, SLAM datasets are often split into several parts to substitute as C-SLAM datasets, as it is, for example, done in \cite{DistributedPGO}.

To address this issue, we present \emph{kollagen}, a collaborative SLAM pose graph generator. 
Inspired by the \Mthreefivehundred{} dataset, \emph{kollagen} generates pose graphs of agents moving randomly on a planar grid. 
For the sake of simplicity, \emph{kollagen} only generates planar pose graphs but it could be easily extended to three-dimensional pose graphs. 
The generator provides both ground truth for the agents, and noisy odometry and loop closure measurements, where the generation of a loop closure is random and the generation probability is based on the proximity of the nodes.
The generated pose graphs are saved in an intuitive file structure for the C-SLAM pose graphs, which lets the user easily extract the pose graph for each agent alone such that it also can be used for the single agent SLAM case. 
Moreover, we give the option to generate the dataset from specifications in a \texttt{.json} file.
A qualitative analysis of the execution time indicates that the pose graph generator scales linearly with the number of agents and the number of nodes of each agent, but scales worse than linear in the radius used for determining loop closures due to the quadratic dependency on the radius.
Furthermore, we demonstrate how to use the pose graph generator and provide two datasets generated with our pose graph generator, which can be readily used as example datasets by the research community to evaluate their algorithms and compare them to other algorithms.

The remainder of the paper is organized as follows. 
In \Cref{sec:RelatedWork}, we review related work concerning pose graph optimization datasets. 
In \Cref{sec:ColeSLAM}, we present \emph{kollagen} and describe its inner workings on a high level, while two example datasets are presented in Section~\ref{sec:BenchmarkDataset}, generated with our pose graph generator and optimized with state-of-the-art pose graph optimizers. 
Finally, \Cref{sec:Conclusions} concludes the paper stating possible extensions of the presented pose graph generator.

\textit{Notation:} The sets of natural and real numbers, and integers are $\mathbb{N}$, $\mathbb{R}$, and $\mathbb{Z}$, respectively. A zero-mean normal distribution with standard deviation $\sigma$ is denoted by $\mathcal{N}(0,\sigma^2)$, while  $\mathcal{U}\{a, b\}$ for integers $a<b$ denotes the discrete uniform distribution over the integers $c$ such that {$a \leq c \leq b$}.
Here, $\wrap: \R \to [-\pi, \pi)$ is a function that
maps an angle into its equivalent angle in the range $[-\pi, \pi)$\footnote{A possible representation of $\wrap$ could be $\theta\mapsto\arctan2(\sin\theta,\cos\theta)$.},
$\%$ is the modulo operator,
and $\delta_i=\delta_{i0}$, where $\delta_{ij}$ is the Kronecker delta.
A planar pose is defined as $\vec x=(x,y,\theta)$, where $x$ and y are the position in the plane, and $\theta$ is the heading. We define $\|\vec x\|_{2,p}$ as the Euclidean norm of the position in the pose, \textit{i.e.}, $\|\vec x\|_{2,p}=\sqrt{x^2+y^2}$.

\section{Related Work}
\label{sec:RelatedWork}
The increased interest in SLAM in the last decades also raised the need for datasets for the proposed SLAM algorithms. 
A classical dataset for planar SLAM is \Mthreefivehundred{}, introduced in \cite{M3500}. 
In this dataset, a mobile agent navigates a grid world, similar to moving around city blocks in Manhattan. 
It consists of 3500 poses and 5600 constraints, which include both odometry and loop closure constraints.
To investigate the robustness to noise of their SLAM algorithm, Carlone \emph{et al.} \cite{Carlone_2014} introduce three variations of \Mthreefivehundred{}, which are based on the same ground truth as \Mthreefivehundred{} but with increased levels of angular noise.
A \Mthreefivehundred{}-like dataset with 10000 nodes (called \texttt{W-10000} in \cite{Grisetti_2010} and \texttt{M10000} in \cite{Carlone_2014_2}) is introduced in \cite{Grisetti_2009} as well as the \sphere{} dataset. 
There also exist many datasets for 3D SLAM algorithms, such as \torus{}, \cube{}, \cubicle{}, and \rim{} introduced in \cite{Carlone_2015}. 
Here, \rim{} and \cubicle{} are pose graphs obtained from real data, while \torus{} and \cube{} are generated via simulation. 
Often, datasets are also published when a new pose graph optimization algorithm for the SLAM back-end is published, as was the case for the datasets \texttt{City10000}, \texttt{CityTrees10000}, and \texttt{sphere2500}, which were released with iSAM \cite{isam}.
A pose graph generator is provided with the g\textsuperscript{2}o framework \cite{g2o}, which is a general graph optimization framework. Their simulator allows for landmark nodes in the graph and different measurement models for the agents. 
However, it requires the g\textsuperscript{2}o framework such that it is not possible to easily integrate it with other pose graph optimizers.

While there exists an abundance of datasets for SLAM, there are not many publicly available datasets for C-SLAM \cite{CollabSLAMSurvey}. 
Nine different datasets with real world data from multi-robot cooperative localization and mapping are presented in \cite{utias}. 
These nine datasets contain the movements of five ground robots and landmark positions.
Five additional real world datasets with three ground, and one aerial, vehicle are provided by \cite{AirMuseumCollabSLAMDataset}. 
These five datasets include IMU measurements and camera images as well as AprilTags for each robot.
Golodetz \emph{et al.} \cite{Collab3DRecon} investigate collaborative 3D reconstruction in four different environments and provide the datasets for these environments.
These C-SLAM datasets concentrate on real world examples, which can be used to evaluate the whole C-SLAM pipeline, \textit{i.e.}, both the front-end and the back-end. 

To obtain datasets for the back-end, it is common to split datasets for single agent SLAM algorithms, such as \Mthreefivehundred{}, \torus{} and \cubicle{}, into
several distinct parts for testing C-SLAM algorithms or to use simple simulations of agents trajectories and loop closures, such as a lawnmower pattern, see, for example, \cite{DistributedPGO}.
However, splitting single SLAM datasets will not necessarily lead to realistic multi-agent datasets, because the obtained trajectories of the agents are often confined to certain areas of the map and do not move as independent agents would potentially move.
One could also use the data provided in \cite{utias,AirMuseumCollabSLAMDataset,Collab3DRecon} to obtain a pose graph, similar to how \cubicle{} and \rim{} are created.
These pose graphs can be useful for benchmarking due to their realistic pose graphs, but for each dataset we would only obtain one pose graph for a certain amount of agents.

Due to the lack of pose graph datasets for C-SLAM, we present, \emph{kollagen}, a data generator for C-SLAM pose graphs, in the following section, which lets the user generate an arbitrary amount of pose graphs efficiently, with an arbitrary amount of agents.

\section{The kollagen Pose Graph Generator}
\label{sec:ColeSLAM}
In this section, we introduce \emph{kollagen}, a collaborative SLAM pose graph generator. 
The \emph{kollagen} data generator is a C++20 header-only library without third-party
dependencies, making it easy to incorporate into existing projects.
It produces a pose graph for each of the $N_{\mathrm{agents}}$ agents as well as inter-agent loop closures connecting the pose graphs of the single agents, where the number of agents, $N_{\mathrm{agents}}\in~\mathbb{N}$, is user-specified.
Furthermore, it offers the ability to output pose graphs in the \texttt{.g2o} file format for use with popular graph optimizing frameworks such as, g\textsuperscript{2}o \cite{g2o} and GTSAM \cite{gtsam}, and ground truth in the \texttt{.tum} file format\footnote{See \textit{Ground-truth trajectories} section at \url{https://vision.in.tum.de/data/datasets/rgbd-dataset/file_formats}} for use with SLAM evaluation tools such as evo \cite{evo}.
However, since the \texttt{.g2o} format does not extend naturally to the multi-agent case, we also present the \textit{multi-g2o} structure  (\Cref{subsec:multig2o}) along with a parser for reading and writing to said structure.

Next, we describe how the collaborative pose graphs are generated, followed by an introduction to our new structure, \textit{multi-g2o}, for saving the collaborative pose graphs, a section on the interoperability of \emph{kollagen}, and finally a qualitative analysis of execution time of \emph{kollagen}. 

\begin{remark}
Please note that in this section, we describe the inner workings of \emph{kollagen} on a high level. If the reader desires more insights on how to use \emph{kollagen}, we invite them to look at the documentation provided on our GitHub page: \url{https://github.com/EricssonResearch/kollagen}
\end{remark}

\subsection{Data Generation}
\begin{figure}[htpb]
    \centering
    \includegraphics[width=1.0\linewidth]{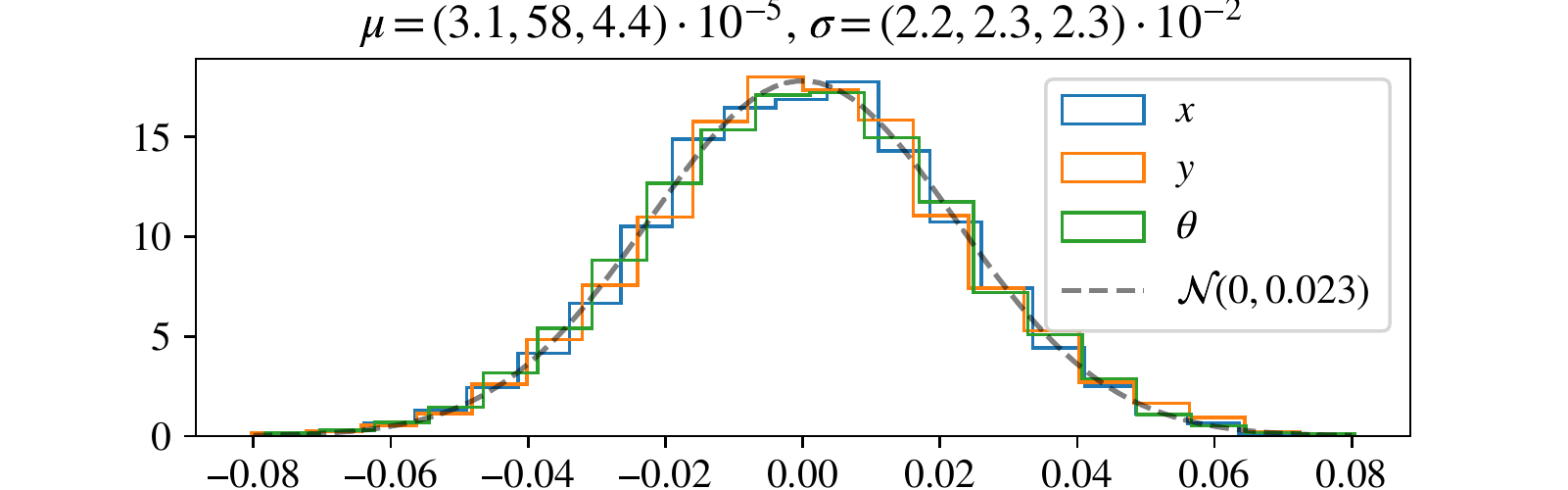}
    \caption{Distribution of the noise influencing the position, $(x,y)$, and the heading, $\theta$, in the \Mthreefivehundred{} dataset}
    \label{fig:M3500statistics}
\end{figure}

The data generator is inspired by the \Mthreefivehundred{} dataset, and we,
therefore, initiate this section with a quick review of \Mthreefivehundred{}.
The pose graph is generated by performing a random walk on a planar grid. 
The agent randomly chooses a direction and then takes four steps in that direction before choosing a new direction. 
This is done in an iterative manner to produce the ground truth.
By taking the ground truth of \Mthreefivehundred{} and comparing the relative translation between each node with the provided odometry measurements, we find that the measured position and heading are seemingly influenced by a zero-mean normal distribution with a standard deviation of $0.023$ as depicted in \Cref{fig:M3500statistics}.
\subsubsection{Agent pose propagation}
Inspired by the \Mthreefivehundred{} \cite{M3500} dataset, the agents are
constricted to moving within a planar Manhattan-type world, that is, a grid of
horizontal and vertical lines with equal spacing. 
Each agent's $k$:th pose, $\vec
x_k=(x_k,y_k,\theta_k)$, is defined by its position $(x_k, y_k)\in\Z^2$ in the $xy$-plane, and its heading angle $\theta_k\in\frac{\pi}{2}\cdot\{-2, -1, 0, 1\}$. 
To generate the ground truth, the agents' poses are propagated as a two-dimensional random walk for $N_{\mathrm{steps}}$ steps, where $N_{\mathrm{steps}}\in\mathbb{N}$ is defined by the user.

Each agent turns in a randomly selected direction (including not turning at all)
before taking $s$ steps forward. 
The parameter $s$ is, by default, the same for all agents and was added taking inspiration from \Mthreefivehundred{} in which
four steps are taken (corresponding to $s=4$) after possibly turning. 
From this description follows that each pose $\vec
x[k]$ can be described as a function of the previous
pose as follows
\begin{subequations}
\begin{align}
\theta_k &= \wrap\left(\theta_{k-1} + \delta_{k \% s}\cdot\frac{\pi}{2}\Theta_{k}\right), \label{eq:stateupdate1} \\
x_k &= x_{k-1} + \cos\theta_k \label{eq:stateupdate2}, \\
y_k &= y_{k-1} + \sin\theta_k \label{eq:stateupdate3},
\end{align}\label{eq:statepropagation}
\end{subequations}
where ${\Theta\sim\mathcal{U}\lbrace-2+n_{d},1\rbrace}$ for a user-defined parameter $n_d\in\lbrace0,1\rbrace$, which enables the possibility of an agent to perform a $180^\circ$ turn when randomly choosing a direction, and $\delta_{k \% s}$ guarantees that turning can only happen every $s$:th step.
The initial pose, $\vec x_0$, of each agent is by default set to zero, but can also be adjusted by the user. 
\begin{remark}
    In \emph{kollagen}, the ground truth is generated on an integer grid according to \Cref{eq:statepropagation} for numerical reasons. However, the results get re-scaled by a factor $\frac{d}{s}$ during saving, where $d\in \mathbb{R}_{>0}$ is specified by the user.
\end{remark}

\subsubsection{Odometry measurement}
In the previous section, we presented the ground truth
pose propagation of an agent given by \Cref{eq:statepropagation}. 
In this section, we describe how an agent takes relative measurements between consecutive poses.

What we expect to obtain from a real system would be noisy measurements of the change in angle
\begin{align}
    \Delta\hat{\theta}_k &= \theta_k-\theta_{k-1} + \tilde{\theta}_k \label{eq:Deltathetaplusone}
\end{align}
and the distance travelled
$\Delta\hat{\ell}_k = \norm{\vec x_k- \vec x_{k-1}}_{2,p} + \tilde{\ell}_k$,
where $\tilde{\theta}\sim\mathcal{N}(0, \sigmaAng^2)$ and $\tilde{\ell}\sim\mathcal{N}(0, \sigmaPos^2)$.
Here, $\sigmaPos$ and $\sigmaAng$ are the (user-defined) standard
deviations of the position and angle, respectively.
This then leads to an estimate of the relative change in $xy$-coordinates given by
\begin{subequations}
\label{eq:RelativeCoordinateChange}
\begin{align}
    \Delta\hat{x}_k &= \Delta\hat{\ell}_k\cos{\Delta\hat{\theta}_k}, \label{eq:xkplusone} \\
    \Delta\hat{y}_k &= \Delta\hat{\ell}_k\sin{\Delta\hat{\theta}_k}, \label{eq:ykplusone}
\end{align}
\end{subequations}
Note that the measurement model described in 
\Cref{,eq:Deltathetaplusone,eq:RelativeCoordinateChange} is biased and has a correlated covariance matrix.

\subsubsection{Pose graph alignment}
Since each agent performs a random walk according to \Cref{eq:statepropagation}, it is likely that the agents' trajectories diverge due to the randomness in each agent's movement.
Depending on the collaborative SLAM application, this can be considered unrealistic behavior, for example, in a constrained environment.
Therefore, we provide an (optional) alignment step which increases the proximity of the agent poses.
In this step, an anchorpoint corresponding to the arithmetic mean of the positions of each agent is calculated. 
Setting the first agent as a reference point, the pose graphs of all the other agents are translated such that their anchorpoints are aligned with the anchorpoint of the first agent.

\subsubsection{Loop closures}
Once the ground truth trajectory and odometry measurments are generated, and the pose graphs have potentially been aligned, \emph{kollagen} generates loop closures as follows.

For each pose $\vec x\in\mathcal{P}_1$, we check whether there exists a pose $\vec z\in \mathcal{P}_2$ such that $\|\vec x - \vec z\|_{2,p}\leq R_{\mathrm{lc}}$, where $R_{\mathrm{lc}}\geq 0$.
If $\|\vec x - \vec z\|_{2,p}=0$, then a loop closure between $\vec x$ and $\vec z$ is added with probability ${p_\mathrm{lc}\in[0,1]}$.
If ${0<\|\vec x - \vec z\|_{2,p}\leq R_{\mathrm{lc}}}$, then a loop closure between $\vec x$ and $\vec z$ is added with a probability proportional to a function that has its maximum at $\|\vec x - \vec z\|_{2,p}=0$ and decreases with an increasing $\|\vec x - \vec z\|_{2,p}$. 
By default the function is set to $\exp\left(-5\left(\|\vec x -\vec z\|_{2,p}/R_\mathrm{lc}\right)^2\right)$.

To generate \emph{intra-agent} loop closures we choose $\mathcal{P}_1=\lbrace\vec x_k\rbrace$ and $\mathcal{P}_2=\lbrace\vec x_i\rbrace_{i=0}^{k-1}$, where $\vec x_i$ is the $i$:th pose of one agent.
Then, for all agents, we iterate through the procedure described above for all $k\in\lbrace 1,\ldots, N_{\mathrm{steps}} \rbrace$ to obtain the intra-loop closures for each agent.

\emph{Inter-agent} loop closures are performed similarly to the intra-agent loop closures, but they are performed pairwise between the agents for the
$N_\mathrm{agents} \choose 2$ possible agent pairs, where $\mathcal{P}_1$ is then the set of all poses of one and $\mathcal{P}_2$ of the other agent, respectively.
Note that both $p_{\mathrm{lc}}$ and $R_{\mathrm{lc}}$ do not have to be the same for intra- and inter-agent loop closures.

The measurement model for the loop closures is based on the true relative change in the coordinates and heading with additive noise. 
For example, for an intra-agent LC let $\vec x_i\in \mathcal{P}_1$ and $\vec x_j\in \mathcal{P}_2$. Then the measurement is given by
\begin{subequations}
\label{eq:RelativeMeasurementsLC}
\begin{align}
    \Delta\hat{\theta}_{ij} &= \theta_i-\theta_{j} + \tilde{\theta}_{ij} \label{eq:Deltathetasimple}\\
    \begin{bmatrix}
        \Delta\hat{x}_{ij} \\
        \Delta\hat{y}_{ij}
    \end{bmatrix} &= 
    \begin{bmatrix}
           \cos\theta_{j} & \sin\theta_{j}\\
           -\sin\theta_{j} & \cos\theta_{j}
    \end{bmatrix}
    \begin{bmatrix}
        x_i-x_{j}\\
        y_i-y_{j}
    \end{bmatrix}+\begin{bmatrix}
        \tilde{x}_{ij}\\
        \tilde{y}_{ij}
    \end{bmatrix} \label{eq:Deltaxysimple},
\end{align}
\end{subequations}
where $\tilde{\theta}\sim\mathcal{N}(0, \sigmaAng^2)$, $\tilde{x}\sim\mathcal{N}(0, \sigmaPos^2)$, $\tilde{y}\sim\mathcal{N}(0, \sigmaPos^2)$, and
$\sigmaPos$ and $\sigmaAng$ are again the (user-defined) standard
deviations of the position and angle for the loop closure, respectively. 
The measurement for a inter-agent loop-closure is defined in a similar manner.
\begin{remark}
We choose a different measurement model for the loop closures than the one in \Cref{eq:Deltathetaplusone,eq:RelativeCoordinateChange}, because the loop closures can be generated with different sensing modalities, which could results in more information than the distance travelled and the angle changed.
Furthermore, note that in contrast to \Cref{eq:Deltathetaplusone,eq:RelativeCoordinateChange} the measurement model in \Cref{eq:RelativeMeasurementsLC} is unbiased and has a diagonal covariance matrix.
\end{remark}

\subsubsection{Generation of the information matrices}
\label{sec:InformationMatrixGeneration}
In a pose graph, each measurement has an (Fisher) information matrix $I$ attached to it, which specifies how much information is contained in the measurement. 
The information matrix can also be interpreted as the inverse of the covariance matrix.
For the odometry, the information matrix is correlated due to the measurement model given by \Cref{eq:Deltathetaplusone,eq:RelativeCoordinateChange}. 
If an agent moves in $x$-direction, the information matrix has correlated $\theta$ and $y$ entries and similarly for a move in $y$-direction.
For loop closures, we have $I=\mathrm{diag}(\sigmaPos^{-2},\sigmaPos^{-2},\sigmaAng^{-2})$ due to the measurement model given by \Cref{eq:RelativeMeasurementsLC}.
Furthermore, we also give the user the ability to choose incorrect diagonal information matrix for both the odometry and loop closures, which can be used to evaluate how their pose graph optimization algorithms handle errors in the information matrix.

\subsection{The multi-g2o structure}\label{subsec:multig2o}
Here, we will introduce the new structure, \textit{multi-g2o}, to save collaborative pose graphs, but first we review the \texttt{.g2o} format and point out its drawbacks for collaborative pose graphs.

The \texttt{.g2o} format is a plaintext format which consists of vertices
(initial guesses), and edges (odometry and loop closures). To take the 2D
planar case as an example, a \texttt{.g2o} file would consist of lines of
\begin{enumerate}[i)]
\item vertices of the form:
\newline\noindent\texttt{VERTEX\_SE2 K x\_K y\_K yaw\_K}\newline
for an identifier \texttt{K}$\in\N$, and (absolute) position/heading estimates given as floating point numbers
\texttt{x\_K}, \texttt{y\_K}, and \texttt{yaw\_K}.
\item edges of the form:
\newline\noindent\texttt{EDGE\_SE2 K\_A K\_B Dx\_K Dy\_K Dyaw\_K I}\newline
which describes the relative transformation \texttt{Dx\_K}, \texttt{Dy\_K}, and
\texttt{Dyaw\_K} from the vertex with identifier \texttt{K\_A} to the vertex with identifier
\texttt{K\_B}.
The entry \texttt{I} denotes the estimated information matrix for the transformation and is given as the upper triangular entries in row-major order. In the 2D case this would hence be given by the six
entries:
\newline\noindent\texttt{I\_11 I\_12 I\_13 I\_22 I\_23 I\_33}\newline
\end{enumerate}
It could be argued that the \texttt{.g2o} format could easily be extended to
multi-agent data by simply allocating a range of identifiers to each agent. The downside to this approach is that the identifiers given to each agent are not apparent within the \texttt{.g2o} file itself, and it is therefore up to the
parser of the file to make this connection. To this end, we propose the
\textit{multi-g2o} structure as illustrated in Listing~\ref{lst:multig2o},
\begin{figure}[htpb]
\begin{minipage}{0.98\linewidth}
\begin{lstlisting}[
  caption=Multi-g2o structure,
  label=lst:multig2o,
]
MultiG2oFolder
|-- inter_agent_lc.dat
|-- agent1
|   |-- posegraph.g2o
|   |-- agent1_GT.tum
|-- agent2
|   |-- posegraph.g2o
|   |-- ...
|-- agent...
|   ...
\end{lstlisting}
\end{minipage}
\end{figure}
where \texttt{MultiG2oFolder} is a directory containing
\begin{itemize}
\item \texttt{inter\_agent\_lc.dat}: 
a plaintext file with rows of similar
      structure as the edges in a \texttt{.g2o} file, that is,
      \newline\noindent\texttt{A1 K1 A2 K2 Dx\_K Dy\_K Dyaw\_K I},\newline
      where \texttt{A1} and \texttt{A2} indicate the agents connected by the inter-agent loop closures, while \texttt{K1} and \texttt{K2} are the identifiers of the connected nodes of the respective agents. 
\item one folder for each agent (here \texttt{agent1}, \texttt{agent2} etc.)
      containing the agent's pose graph with intra-agent loop closures as a
      \texttt{.g2o} file, and the ground truth of the agent in \texttt{.tum} file
      format (necessary for some SLAM evaluation frameworks).
\end{itemize}
This structure has the following benefits:
\begin{enumerate}[1.]
\item It clearly distinguishes data for each agent: it is possible to use the \texttt{.g2o} file of each agent separately in, \eg, GTSAM,
      g\textsuperscript{2}o, etc.;
\item It is human-readable and easily navigable through the OS file browser;
\item The provided ground truth \texttt{.tum} file makes it easy to use with the
      popular evaluation tool evo \cite{evo}.
\end{enumerate}

\subsection{Interoperability}\label{subsec:interoperability}

The \emph{kollagen} library can be used in a number of ways depending on where and how the data is meant to be used. 
The highest number of possibilities is given to users targeting C++:
these users can include the library header \texttt{kollagen.h}, from which they
can generate data for immediate use, without any need of saving to disk. An example of this
is provided in \texttt{iSAM2example.cpp} where data is generated and converted to GTSAM types,
allowing direct usage with, \eg, iSAM2. If the end target is not C++, a more convenient option
is to generate data by building and running \texttt{generate.cpp} or to install the python package using pip.
The generated data can then be saved as a single \texttt{.g2o} file which then needs to be
parsed (no parser provided, but see, \eg, \cite{DistributedPGO} for an example), 
or saved as \textit{multi-g2o}, for which a parser for C++ is provided.
In addition to that, the datasets can be generated via specifications given in a \texttt{.json} file, which means that whole datasets can be shared and regenerated via just one single file.

\subsection{Execution time}
\begin{figure*}[t]
    \centering
    \begin{subfigure}[t]{0.32\textwidth}
        \centering
        \includegraphics[trim={0 0 0 1cm},clip,width=1.1\linewidth]{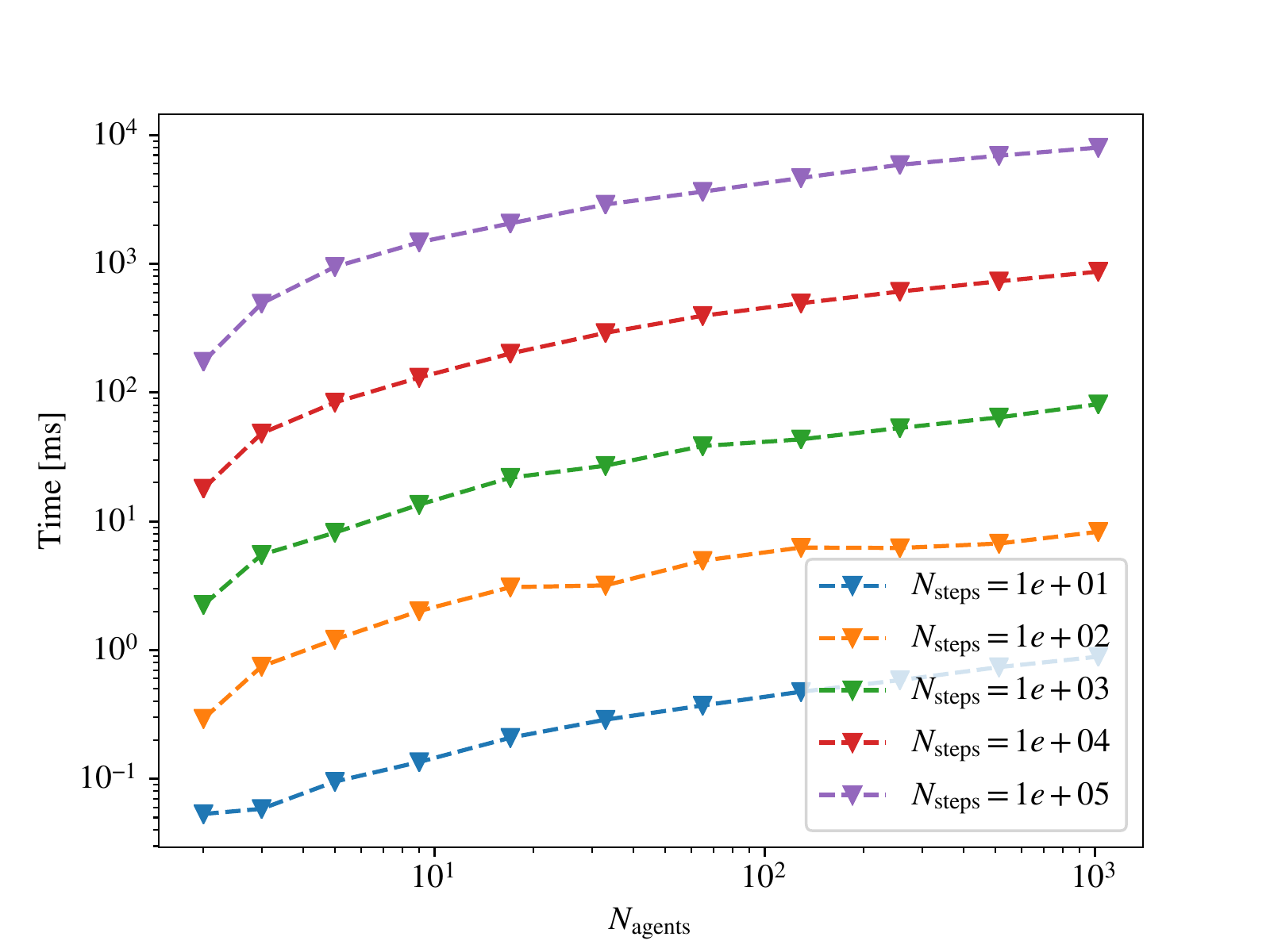}
        \caption{}
        \label{fig:Nagents}
    \end{subfigure}
    \begin{subfigure}[t]{0.32\textwidth}
        \centering
        \includegraphics[trim={0 0 0 1cm},clip,width=1.1\linewidth]{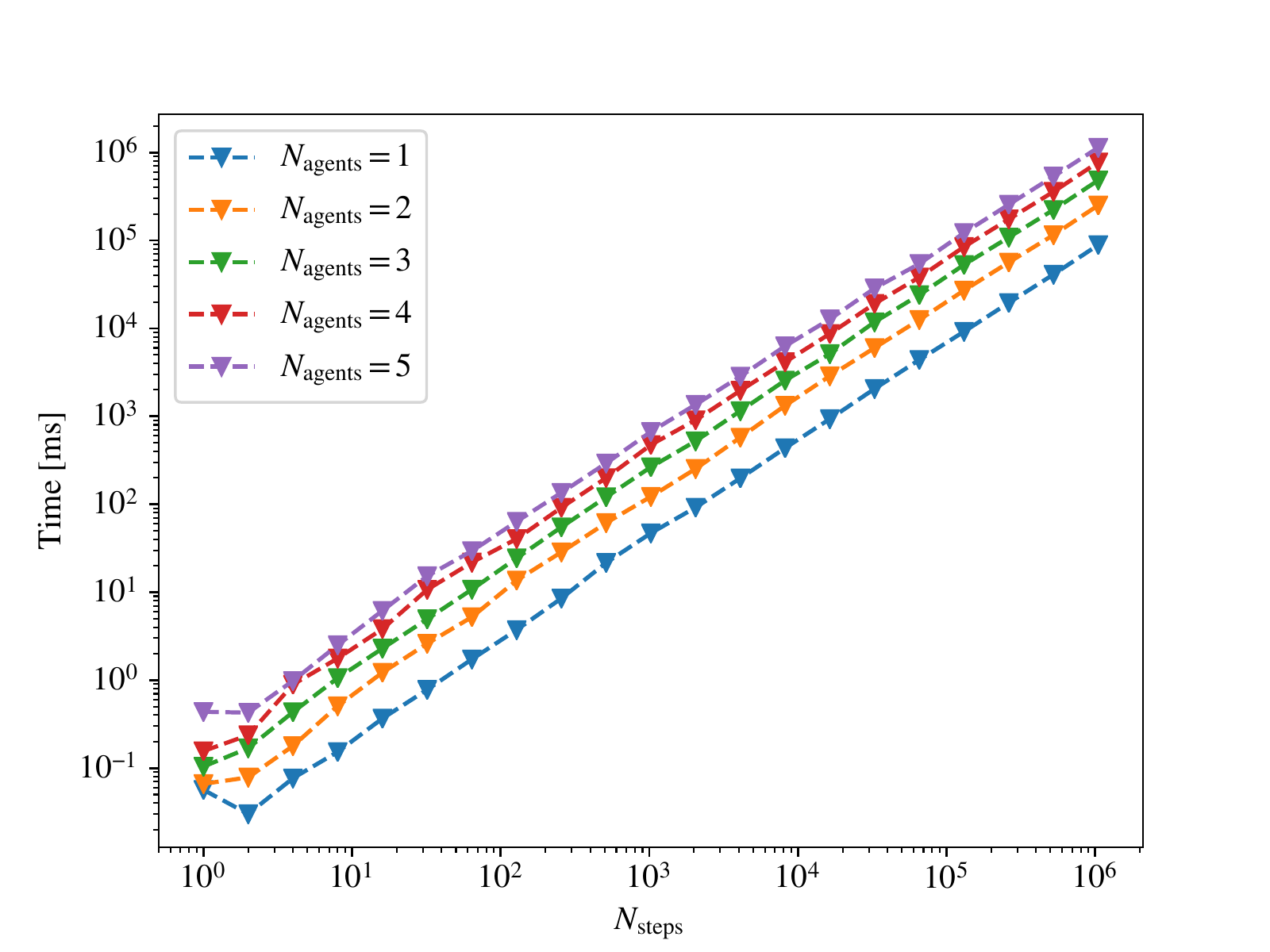}
        \caption{}
        \label{fig:Nstep}
    \end{subfigure}
    \begin{subfigure}[t]{0.32\textwidth}
        \centering
        \includegraphics[trim={0 0 0 1cm},clip,width=1.1\linewidth]{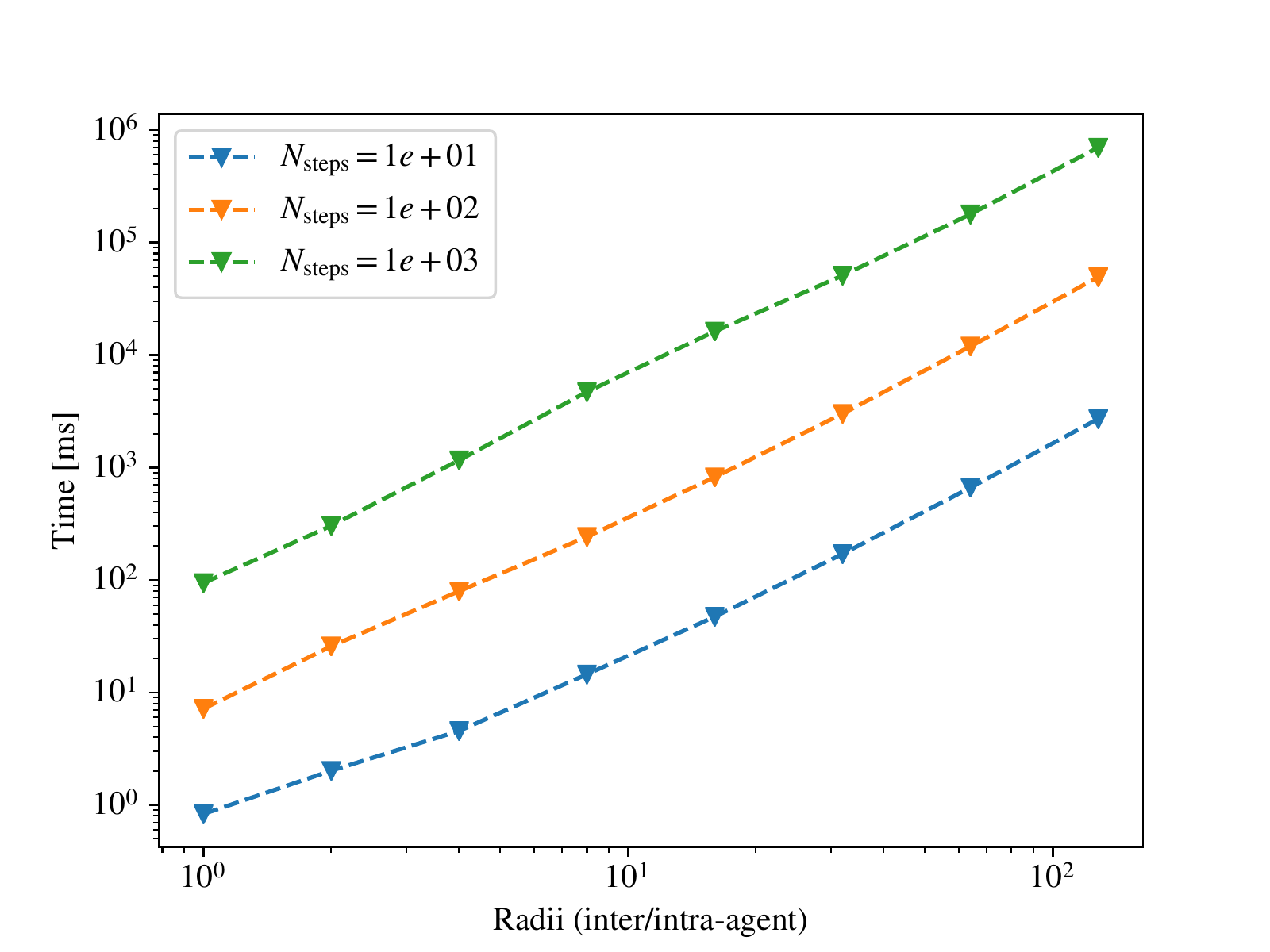}
        \caption{}
        \label{fig:Radius}
    \end{subfigure}
    \caption{(a) Execution time VS the number of agents $N_\mathrm{agents}$, for a selection of values of $N_\mathrm{steps}$; (b) Execution time VS the number of steps $N_\mathrm{steps}$, for a selected number of agents $N_\mathrm{agents}$; (c) Execution time VS the size of the radii, for a selected number of steps $N_\mathrm{steps}$.}
    \label{fig:}
\end{figure*}
While the execution time may not be critical in most use-cases, some users of
\emph{kollagen} might -- for practical reasons -- want the generation to be both fast
and scalable. One such case could be the need for massive data-generation in
learning-based applications. Another might be to generate a large variation of
datasets to get a statistically sound perception of performance or accuracy. 
For this reason we provide \Cref{fig:Nagents,fig:Nstep,fig:Radius} which showcase the
execution times when increasing the number of steps taken, the number of
agents, and the size of the distance used in the intra- and inter-agent loop
closure generation, respectively.
In \Cref{fig:Radius}, we set $R_{\mathrm{lc}}=R$ for both intra- and inter-agent loop closures and vary $R$.
The tests were performed on a laptop with an Intel i7-1185G7 CPU @ 3.0 GHz and 32 GB RAM.

The data suggests that the execution time scales linearly both in terms of the number
of steps and agents. 
However, the data indicates a worse than linear time increase when we increase $R$.
While this might be expected due to the number of look-ups being proportional to $R^2$, we are pointing this out to users that want to work with high values of $R$, since they can expect increased times for
generating datasets. 
For details on the exact parameters and procedures used
for generating these figures, we refer the reader to the source file
\texttt{timing.cpp} in the \emph{kollagen} library.

In general, \emph{kollagen} has been optimized for readability, maintainability, and
ease of use rather than for speed. This means that there are several
close-at-hand code optimizations that users with stricter speed requirements can
perform. One such optimization would be to replace the $\Ordo(\log{N})$ \texttt{std::map}
containers with $\Ordo(1)$ \texttt{std::unordered\_map} containers. However,
since not all \texttt{std::map} containers in \emph{kollagen} are holding types that are
hashable by default (such as \texttt{std::pair}), such a change would require either
implementing hashing functions for such types, or to include 
the Boost C++ Libraries (\url{https://www.boost.org/}) as a dependency and use 
the hashing functions they provide.

\section{Example Datasets}
\label{sec:BenchmarkDataset}
In this section, we use the data generator to provide two example
datasets generated with \emph{kollagen} that can be used by the community out-of-the-box to evaluate their pose
graph optimization algorithms for collaborative SLAM. 
We first describe the datasets and then apply three state-of-the-art algorithms on these example datasets to showcase that our generated datasets can be processed by state-of-the-art algorithms.

Let us now describe the two example datasets that are provided with
this paper and are generated with \emph{kollagen}. 
\begin{figure}[htpb]
    \begin{subfigure}[t]{0.56\columnwidth}
        \centering
        \includegraphics[height=4.3cm]{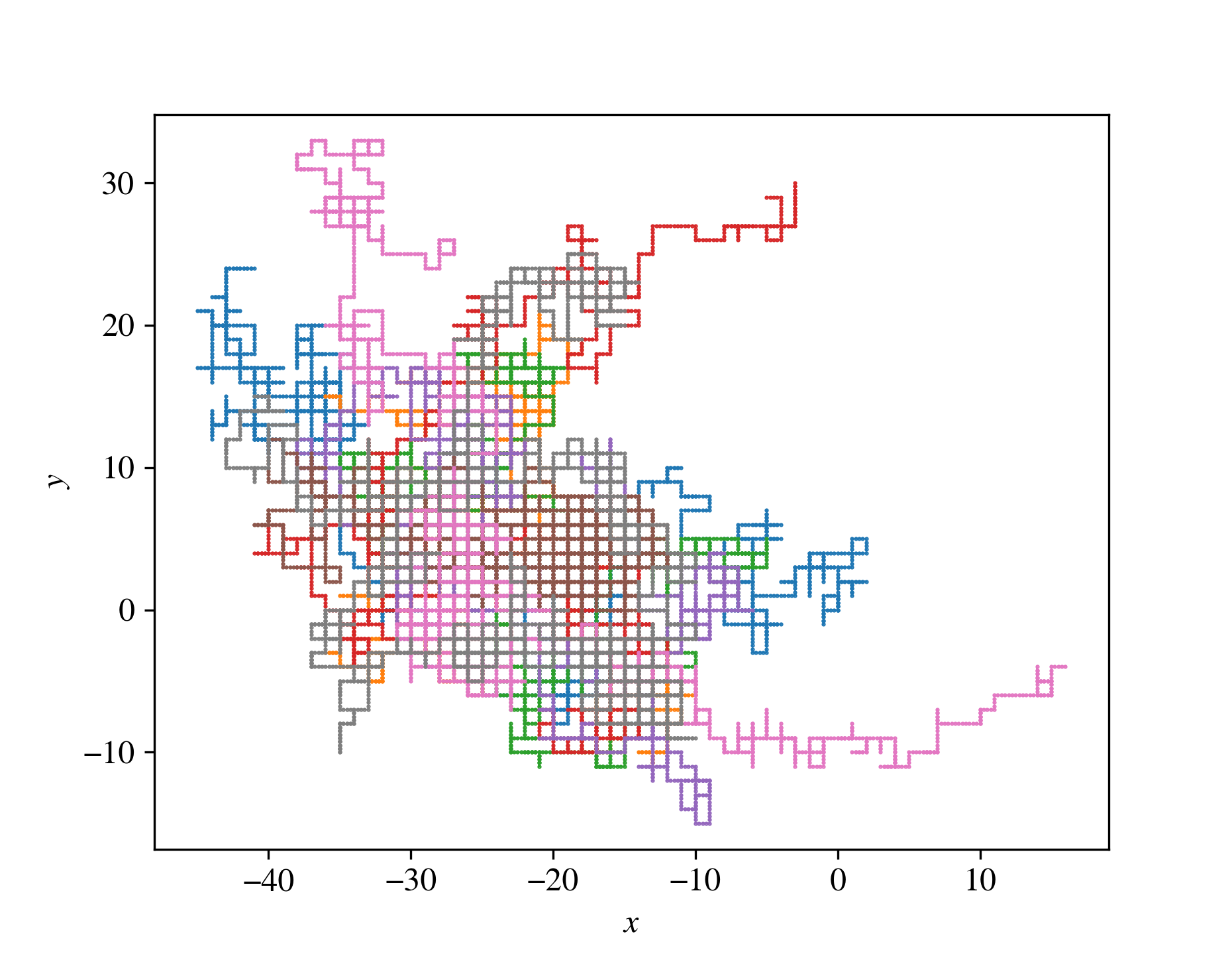}
        \caption{}
        \label{fig:TrajectoriesOfMulti3500x8}
    \end{subfigure}
    \begin{subfigure}[t]{0.43\columnwidth}
        \centering
        \includegraphics[height=4.3cm]{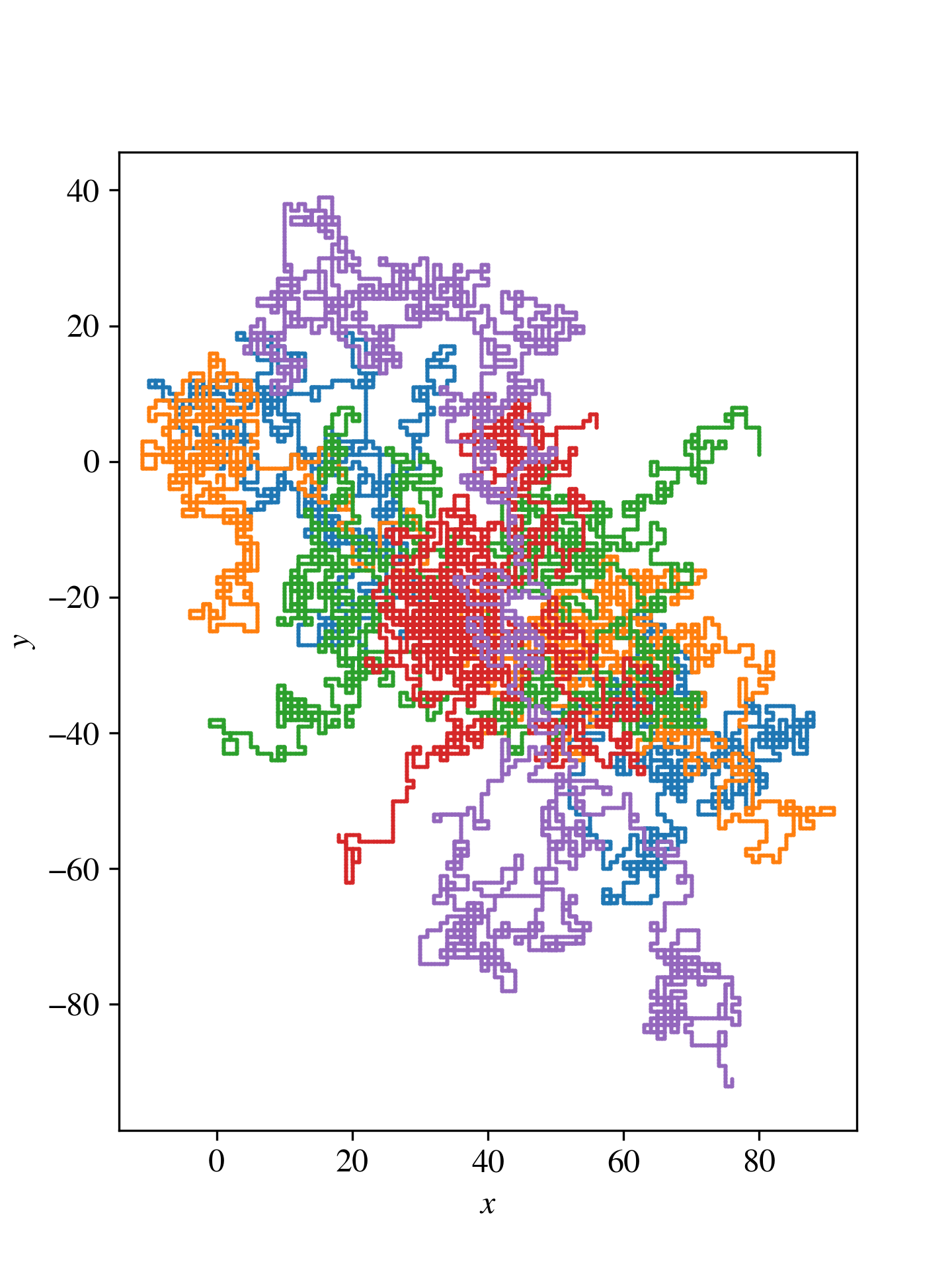}
        \caption{}
        \label{fig:TrajectoriesOfMulti10000x5}
    \end{subfigure}
    \caption{Ground truth trajectories of (a) the eight agents in \MultiThreeFivehundredXeight{}; (b) the five agents in \MultiTenThousandXfive{}.}
    \label{fig:benchmark-datasets}
\end{figure}
The first dataset, \MultiThreeFivehundredXeight{}, is inspired by
\Mthreefivehundred{} and consists of eight agents, which have 3500 nodes in
their respective factor graphs.
Since the agent in \Mthreefivehundred{} is able to perform $180^\circ$ turns, we also enable that option for this dataset (as described below \Cref{eq:statepropagation}). 
In total, the dataset has 67645 constraints, where 39645 out of these constraints are loop closures.
The ground truth trajectories of the eight agents are shown in \Cref{fig:TrajectoriesOfMulti3500x8}.
The parameter
settings used to generate \MultiThreeFivehundredXeight{} can be found in \MultiThreeFivehundredXeight{}\texttt{.json}

The second dataset, \MultiTenThousandXfive{}, is inspired by the \texttt{M10000} dataset and consists of five agents, which have 10000 nodes in their respective factor graphs. 
In this dataset, we disabled the $180^\circ$ turns of the agents such that they can only turn left and right, or not keep the same direction.
This dataset has a total of 76134 constraints out of which 26134 are loop closure constraints.
The ground truth trajectories of the five agents are shown in
\Cref{fig:TrajectoriesOfMulti10000x5}. We foresee that the use of this
dataset is to evaluate the performance of collaborative pose graph optimizers
for large datasets, \eg, one can analyze the speed of the pose graph optimizer.
The parameter settings used to generate \MultiTenThousandXfive{} can be found in \MultiTenThousandXfive{}\texttt{.json}

Next, we will run state-of-the-art algorithms for pose graph optimization on 
our two example datasets. 
The state-of-the-art algorithms we run our example datasets on are SE-Sync~\cite{se-sync}, DPGO~\cite{DistributedPGO}, and g\textsuperscript{2}o~\cite{g2o}.
For SE-Sync we use the default parameters with chordal initialization and we let DPGO run for 50 iterations with default parameters. For g\textsuperscript{2}o we initialize with \textit{spanning tree} and use the \texttt{gn\_var\_cholmod} solver with default parameters for 10 iterations.
SE-Sync, DPGO, and g\textsuperscript{2}o expect a single \texttt{.g2o} file as an input. Therefore, we need to concatenate all agent graphs for both investigated datasets in one \texttt{.g2o} file, which is handled by \emph{kollagen}. 
These \texttt{.g2o} files can then be directly processed in SE-Sync or g\textsuperscript{2}o. For DPGO, we use the accompanying \texttt{MultiRobotExample.cpp}.

Since we give the user the option to save the \texttt{.g2o} files with the correct  information matrices and incorrect diagonal matrices (see Section~\ref{sec:InformationMatrixGeneration}), we use both correct and incorrect information matrices for the odometry, where the incorrect matrices are given by $I=\mathrm{diag}(\sigmaPos^{-2},\sigmaPos^{-2},\sigmaAng^{-2})$. 
The loop closures have the correct information matrices though.
We call the case with incorrect diagonal information matrices \texttt{Diagonal} and the case with the correct information matrices \texttt{Exact}.
Since the objective functions of the optimizers typically depend on the information matrices, we cannot simply compare the optimal values of the objective values in the two considered cases.
To obtain a fair comparison of the pose estimates, we will use the unaligned mean average pose error (APE) with respect to the translations, as determined by evo \cite{evo}, between the ground truth and the pose estimates.

The results for \MultiThreeFivehundredXeight{} and \MultiTenThousandXfive{} are shown in Table~\ref{tab:M3500x8Eval} and Table~\ref{tab:M10000x5Eval}, respectively. 
First, we note each of the pose graph optimizers has reduced the mean APE significantly compared to the mean APE obtained from the odometry estimates. This shows that the state-of-the-art optimizers have sucessfully used our example datasets to generate a better estimate of the poses than the odometry.
For SE-Sync and DPGO, we observe that if the diagonal information matrices are used for the odometry the mean APE increases compared to using the exact information matrix, which is an expected outcome. 
Furthermore, we see that SE-Sync outperforms DPGO, which is again expected since SE-Sync is a global solver and we have limited DPGO to 50 iterations.
However, g\textsuperscript{2}o behaves in an interesting way in the diagonal case. For \MultiThreeFivehundredXeight{}, the obtained mean APE for g\textsuperscript{2}o is even smaller than the one obtained for SE-Sync and DPGO. 
While SE-Sync has returned the globally optimal pose estimates for its objective function, we would like to point out that g\textsuperscript{2}o solves a different objective value, which could be one explanation why the estimates of g\textsuperscript{2}o are closer to the ground truth than the estimates of SE-Sync.
Furthermore, returning the globally optimal solution to an objective function does not imply that the obtained solution will be close to the ground truth, since the objective function is influenced by the noisy measurements, as, \eg, pointed out in \cite{SpectralInitialization}.
In the case with the true information matrices, SE-Sync performs better than g\textsuperscript{2}o, which can be explained with the fact that SE-Sync now has the correct information matrices in its objective function such that it will get closer to the ground truth.

\begin{table}
\caption{Results for the mean APE on \MultiThreeFivehundredXeight{}}
\label{tab:M3500x8Eval}
\centering
\begin{tabular}{|c|c|c|c|c|}
\hline 
 Case & g\textsuperscript{2}o & DPGO & SE-Sync & Odometry \\ 
\hline 
\texttt{Diagonal}  & 0.385161 & 0.966504 & 0.733618 & 14.883697 \\ 
\hline 
\texttt{Exact} & 0.390065 & 0.252501 & 0.212746 & 14.883697 \\ 
\hline 
\end{tabular} 
\end{table}

\begin{table}
\caption{Results for the mean APE on \MultiTenThousandXfive{}}
\label{tab:M10000x5Eval}
\centering
\begin{tabular}{|c|c|c|c|c|}
\hline 
 Case & g\textsuperscript{2}o & DPGO & SE-Sync & Odometry \\ 
\hline 
\texttt{Diagonal}  & 1.019478 & 3.098239 & 0.791197 & 17.427713 \\ 
\hline 
\texttt{Exact} & 1.426272 & 1.178893 & 0.446346 & 17.427713 \\ 
\hline 
\end{tabular} 
\end{table}

\section{Conclusions}
\label{sec:Conclusions}
In this paper, we presented \emph{kollagen}, a collaborative SLAM pose graph generator. 
\emph{kollagen} provides a simple and intuitive way to generate pose graphs in a collaborative SLAM setting, which has been lacking so far. 
Therefore, it provides the means to generate datasets for collaborative SLAM pose graph optimizers, which will enhance both the reproducibility of results as well as the comparison of different optimizers with high quality datasets.

The current version of \emph{kollagen} produces a random walk on a planar grid for a user-specified amount of agents and also allows the user to specify the number of nodes for the agents' pose graphs. 
The odometry and loop closure measurements are influenced by normally distributed noise processes with user-defined covariances, while the probability of creating a loop closure can be adjusted by the user.
Finally, we also showcased how state-of-the-art pose graph optimization algorithms can run on two provided datasets.

There are several possible extensions that could be made to \emph{kollagen}.
First, the current version produces only planar pose graphs, where the robot at each time step either moves straight ahead, left, right, or backwards. 
A simple extension for the three dimensional case is to let the robot also move up and down randomly. 
This would lead to a similar dataset as \cube{} but with several agents.
Second, the odometry and loop closure measurements generated are based on normally distributed noise. One extension could be to use a different noise distribution to have more realistic measurements, while another extension could be to modify the code to randomly introduce outliers.
\newpage

{
\bibliographystyle{IEEEtran}  
\bibliography{biblio}  

\begin{thebibliography}{10}
\providecommand{\url}[1]{#1}
\csname url@samestyle\endcsname
\providecommand{\newblock}{\relax}
\providecommand{\bibinfo}[2]{#2}
\providecommand{\BIBentrySTDinterwordspacing}{\spaceskip=0pt\relax}
\providecommand{\BIBentryALTinterwordstretchfactor}{4}
\providecommand{\BIBentryALTinterwordspacing}{\spaceskip=\fontdimen2\font plus
\BIBentryALTinterwordstretchfactor\fontdimen3\font minus
  \fontdimen4\font\relax}
\providecommand{\BIBforeignlanguage}[2]{{%
\expandafter\ifx\csname l@#1\endcsname\relax
\typeout{** WARNING: IEEEtran.bst: No hyphenation pattern has been}%
\typeout{** loaded for the language `#1'. Using the pattern for}%
\typeout{** the default language instead.}%
\else
\language=\csname l@#1\endcsname
\fi
#2}}
\providecommand{\BIBdecl}{\relax}
\BIBdecl

\bibitem{SLAMSurvey}
\BIBentryALTinterwordspacing
C.~Cadena, L.~Carlone, H.~Carrillo, Y.~Latif, D.~Scaramuzza, J.~Neira, I.~Reid,
  and J.~J. Leonard, ``Past, present, and future of simultaneous localization
  and mapping: Toward the robust-perception age,'' \emph{{IEEE} Transactions on
  Robotics}, vol.~32, no.~6, pp. 1309--1332, Dec 2016. [Online]. Available:
  \url{https://doi.org/10.1109%2Ftro.2016.2624754}
\BIBentrySTDinterwordspacing

\bibitem{CollabSLAMSurvey}
\BIBentryALTinterwordspacing
P.-Y. Lajoie, B.~Ramtoula, F.~Wu, and G.~Beltrame, ``Towards collaborative
  simultaneous localization and mapping: a survey of the current research
  landscape,'' \emph{Field Robotics}, vol.~2, no.~1, pp. 971--1000, Mar 2022.
  [Online]. Available: \url{https://doi.org/10.55417%2Ffr.2022032}
\BIBentrySTDinterwordspacing

\bibitem{iSAM2}
\BIBentryALTinterwordspacing
M.~Kaess, H.~Johannsson, R.~Roberts, V.~Ila, J.~Leonard, and F.~Dellaert,
  ``{iSAM}2: Incremental smoothing and mapping with fluid relinearization and
  incremental variable reordering,'' in \emph{2011 {IEEE} International
  Conference on Robotics and Automation}.\hskip 1em plus 0.5em minus
  0.4em\relax {IEEE}, May 2011. [Online]. Available:
  \url{https://doi.org/10.1109%2Ficra.2011.5979641}
\BIBentrySTDinterwordspacing

\bibitem{se-sync}
\BIBentryALTinterwordspacing
D.~M. Rosen, L.~Carlone, A.~S. Bandeira, and J.~J. Leonard, ``{SE}-{S}ync: A
  certifiably correct algorithm for synchronization over the special euclidean
  group,'' \emph{The International Journal of Robotics Research}, vol.~38, no.
  2-3, pp. 95--125, Aug 2018. [Online]. Available:
  \url{https://doi.org/10.1177%2F0278364918784361}
\BIBentrySTDinterwordspacing

\bibitem{M3500}
\BIBentryALTinterwordspacing
E.~Olson, J.~Leonard, and S.~Teller, ``Fast iterative alignment of pose graphs
  with poor initial estimates,'' in \emph{Proceedings 2006 {IEEE} International
  Conference on Robotics and Automation, 2006. {ICRA} 2006.}\hskip 1em plus
  0.5em minus 0.4em\relax {IEEE}. [Online]. Available:
  \url{https://doi.org/10.1109%2Frobot.2006.1642040}
\BIBentrySTDinterwordspacing

\bibitem{Carlone_2015}
\BIBentryALTinterwordspacing
L.~Carlone, R.~Tron, K.~Daniilidis, and F.~Dellaert, ``Initialization
  techniques for 3d {SLAM}: A survey on rotation estimation and its use in pose
  graph optimization,'' in \emph{2015 {IEEE} International Conference on
  Robotics and Automation ({ICRA})}.\hskip 1em plus 0.5em minus 0.4em\relax
  {IEEE}, May 2015. [Online]. Available:
  \url{https://doi.org/10.1109%2Ficra.2015.7139836}
\BIBentrySTDinterwordspacing

\bibitem{DistributedPGO}
\BIBentryALTinterwordspacing
Y.~Tian, K.~Khosoussi, D.~M. Rosen, and J.~P. How, ``Distributed certifiably
  correct pose-graph optimization,'' \emph{{IEEE} Transactions on Robotics},
  vol.~37, no.~6, pp. 2137--2156, Dec 2021. [Online]. Available:
  \url{https://doi.org/10.1109%2Ftro.2021.3072346}
\BIBentrySTDinterwordspacing

\bibitem{MajorMinPoseGraph}
T.~Fan and T.~Murphey, ``Majorization minimization methods for distributed pose
  graph optimization with convergence guarantees,'' in \emph{2020 IEEE/RSJ
  International Conference on Intelligent Robots and Systems (IROS)}, 2020, pp.
  5058--5065.

\bibitem{Carlone_2014}
\BIBentryALTinterwordspacing
L.~Carlone and A.~Censi, ``From angular manifolds to the integer lattice:
  Guaranteed orientation estimation with application to pose graph
  optimization,'' \emph{{IEEE} Transactions on Robotics}, vol.~30, no.~2, pp.
  475--492, Apr 2014. [Online]. Available:
  \url{https://doi.org/10.1109%2Ftro.2013.2291626}
\BIBentrySTDinterwordspacing

\bibitem{Grisetti_2010}
\BIBentryALTinterwordspacing
G.~Grisetti, R.~K\"ummerle, C.~Stachniss, U.~Frese, and C.~Hertzberg,
  ``Hierarchical optimization on manifolds for online 2d and 3d mapping,'' in
  \emph{2010 {IEEE} International Conference on Robotics and Automation}.\hskip
  1em plus 0.5em minus 0.4em\relax {IEEE}, May 2010. [Online]. Available:
  \url{https://doi.org/10.1109%2Frobot.2010.5509407}
\BIBentrySTDinterwordspacing

\bibitem{Carlone_2014_2}
\BIBentryALTinterwordspacing
L.~Carlone, R.~Aragues, J.~A. Castellanos, and B.~Bona, ``A fast and accurate
  approximation for planar pose graph optimization,'' \emph{The International
  Journal of Robotics Research}, vol.~33, no.~7, pp. 965--987, May 2014.
  [Online]. Available: \url{https://doi.org/10.1177%2F0278364914523689}
\BIBentrySTDinterwordspacing

\bibitem{Grisetti_2009}
\BIBentryALTinterwordspacing
G.~Grisetti, C.~Stachniss, and W.~Burgard, ``Nonlinear constraint network
  optimization for efficient map learning,'' \emph{{IEEE} Transactions on
  Intelligent Transportation Systems}, vol.~10, no.~3, pp. 428--439, Sep 2009.
  [Online]. Available: \url{https://doi.org/10.1109%2Ftits.2009.2026444}
\BIBentrySTDinterwordspacing

\bibitem{isam}
\BIBentryALTinterwordspacing
M.~Kaess, A.~Ranganathan, and F.~Dellaert, ``{iSAM}: Fast incremental smoothing
  and mapping with efficient data association,'' in \emph{Proceedings 2007
  {IEEE} International Conference on Robotics and Automation}.\hskip 1em plus
  0.5em minus 0.4em\relax {IEEE}, apr 2007. [Online]. Available:
  \url{https://doi.org/10.1109%2Frobot.2007.363563}
\BIBentrySTDinterwordspacing

\bibitem{g2o}
\BIBentryALTinterwordspacing
R.~K\"ummerle, G.~Grisetti, H.~Strasdat, K.~Konolige, and W.~Burgard, ``g2o: A
  general framework for graph optimization,'' in \emph{2011 {IEEE}
  International Conference on Robotics and Automation}.\hskip 1em plus 0.5em
  minus 0.4em\relax {IEEE}, May 2011. [Online]. Available:
  \url{https://doi.org/10.1109%2Ficra.2011.5979949}
\BIBentrySTDinterwordspacing

\bibitem{utias}
\BIBentryALTinterwordspacing
K.~Y. Leung, Y.~Halpern, T.~D. Barfoot, and H.~H. Liu, ``The {UTIAS}
  multi-robot cooperative localization and mapping dataset,'' \emph{The
  International Journal of Robotics Research}, vol.~30, no.~8, pp. 969--974,
  Mar 2011. [Online]. Available:
  \url{https://doi.org/10.1177%2F0278364911398404}
\BIBentrySTDinterwordspacing

\bibitem{AirMuseumCollabSLAMDataset}
\BIBentryALTinterwordspacing
R.~Dubois, A.~Eudes, and V.~Fremont, ``{AirMuseum}: a heterogeneous multi-robot
  dataset for stereo-visual and inertial simultaneous localization and
  mapping,'' in \emph{2020 {IEEE} International Conference on Multisensor
  Fusion and Integration for Intelligent Systems ({MFI})}.\hskip 1em plus 0.5em
  minus 0.4em\relax {IEEE}, sep 2020. [Online]. Available:
  \url{https://doi.org/10.1109%2Fmfi49285.2020.9235257}
\BIBentrySTDinterwordspacing

\bibitem{Collab3DRecon}
\BIBentryALTinterwordspacing
S.~Golodetz, T.~Cavallari, N.~A. Lord, V.~A. Prisacariu, D.~W. Murray, and
  P.~H.~S. Torr, ``Collaborative large-scale dense 3d reconstruction with
  online inter-agent pose optimisation,'' \emph{{IEEE} Transactions on
  Visualization and Computer Graphics}, vol.~24, no.~11, pp. 2895--2905, nov
  2018. [Online]. Available: \url{https://doi.org/10.1109%2Ftvcg.2018.2868533}
\BIBentrySTDinterwordspacing

\bibitem{gtsam}
\BIBentryALTinterwordspacing
F.~Dellaert, R.~Roberts, A.~Cunningham, {Varun Agrawal}, C.~Beall, {Duy-Nguyen
  Ta}, {Lucacarlone}, F.~Jiang, {Nikai}, J.~L. Blanco-Claraco, S.~Williams,
  {Ydjian}, A.~Melim, {Zhaoyang Lv}, J.~Dong, J.~Lambert, {Krunal Chande},
  {Akshay Krishnan}, G.~Chen, {Balderdash-Devil}, {DiffDecisionTrees}, {Sungtae
  An}, {Mpaluri}, {Ellon Paiva Mendes}, M.~Bosse, A.~Patel, {Ayush Baid},
  P.~Furgale, {Matthewbroadwaynavenio}, and {Roderick-Koehle}, ``borglab/gtsam:
  4.1.1,'' 2021. [Online]. Available: \url{https://zenodo.org/record/5794542}
\BIBentrySTDinterwordspacing

\bibitem{evo}
M.~Grupp, ``evo: Python package for the evaluation of odometry and slam.''
  \url{https://github.com/MichaelGrupp/evo}, 2017.

\bibitem{SpectralInitialization}
K.~J. Doherty, D.~M. Rosen, and J.~J. Leonard, ``Performance guarantees for
  spectral initialization in rotation averaging and pose-graph slam,'' in
  \emph{2022 International Conference on Robotics and Automation (ICRA)}, 2022,
  pp. 5608--5614.

\end{thebibliography}
}

\end{document}